\documentclass[11pt]{article}

\usepackage{acl}
\usepackage{times}
\usepackage{latexsym}
\usepackage[T1]{fontenc}
\usepackage[utf8]{inputenc}
\usepackage{microtype}
\usepackage{inconsolata}
\usepackage{graphicx}
\usepackage{tabularx}
\usepackage{booktabs}

\title{cantnlp@DravidianLangTech 2026: organic domain adaptation improves multi-class hope speech detection in Tulu}

\author{Andrew Li \\
  Lake Washington School District \\
  \texttt{landrewi@hotmail.com} \\\And
  Sidney Wong \\
  University of Otago, New Zealand \\
  Te Pūnaha Matatini, New Zealand \\
  \texttt{sidney.wong@otago.ac.nz} \\}

\begin{document}
\maketitle
\begin{abstract}

    This paper presents our systems and results for the Hope Speech Detection in Code-Mixed Tulu Language shared task at the Sixth Workshop on Speech, Vision, and Language Technologies for Dravidian Languages (DravidianLangTech-2026). We trained an XLM-RoBERTa-based text classification system for detecting hope speech in code-mixed Tulu social media comments. We compared this organically adapted hope speech detection model with our baseline model. On the development set, the organically adapted model outperformed the baseline system. While our submitted systems performed more modestly on the official test set, these results suggest that further adapting XLM-RoBERTa on organically collected Tulu social media text containing code-mixed and mixed-script variation can improve hope speech detection in code-mixed Tulu.
    
\end{abstract}

\section{Introduction}
\label{sec:introduction}

    There has been increasing interest in detecting different forms of language on social media, including hate speech, offensive language, and other emotionally meaningful content. More recently, researchers have also begun to study hope speech, which includes supportive, encouraging, and empathetic expressions in online communication (\citealp{chhabra_literature_2023}; \citealp{ghosal_hope_2026}). Detecting hope speech is useful for promoting healthier online discourse, identifying constructive engagement, and supporting research on emotional wellbeing in low-resource communities \cite{chakravarthi_overview_2022}. These goals become more difficult in under-resourced languages, where limited labeled data and strong linguistic variation can reduce the effectiveness of standard text classification systems. This challenge is especially apparent in code-mixed settings, where multiple languages and scripts may appear within the same comment.
    
    In this paper, we describe our system for the Hope Speech Detection in Code-Mixed Tulu shared task at DravidianLangTech-2026 \cite{thenmozhi_findings_2026}. The purpose of this shared task is to develop a multi-level hope speech classification system in code-mixed Tulu. Our approach used XLM-RoBERTa as a transformer-based text classifier \cite{conneau_unsupervised_2020}. Furthermore, we extended a similar fine-tuning strategy used in prior shared task work \cite{wong_cantnlplt-edi-2024_2024}. Our primary motivation was to account for the script-switching phenomena observed in the training data to ensure that we account for the orthographic (and linguistic) context. We further adapt the model using additional organic Tulu social media text containing code-mixed and mixed-script variation to better match the linguistic patterns found in real Tulu social media text.
    
\section{Related Works}
\label{sec:related_works}

    The first shared task to explored hope speech detection in YouTube comments across three language conditions: English, Tamil, and Malayalam \cite{chakravarthi_findings_2021}. In descending order, the best performing hope speech detection models based on weighted average $F_1$-score were for English at 0.93, Malayalam at 0.85 \cite{hossain_nlp-cuetlt-edi-eacl2021_2021}, and Tamil at 0.61 \cite{sharma_spartanslt-edi-eacl2021_2021}. Unsurprisingly, the best performing hope speech detection models utilized variants of the Bidirectional Encoder Representations from Transformers (BERT) model architecture \cite{devlin_bert_2019}. This contrasts earlier detection systems which relied on shallow classifiers \cite{chakravarthi_hope_2022}.
    
    Successive shared tasks focusing on hope speech detection would expand to other language conditions. The second shared task, also with YouTube comments, included Kannada and Spanish \cite{chakravarthi_overview_2022-1}. Meanwhile, the third shared task focused on three Indo-European languages: Bulgarian, English, Hindi, and Spanish \cite{kumaresan_overview_2023}. Using data from X, formerly known as Twitter, the best performing model for English achieved a weighted average $F_1$-score of 0.50 using shallow classifiers (\citealt{kumari_mlai_iiitranchilt-edi-2023_2023}; \citealt{thandavamurthi_tercetlt-edi-2023_2023}). Related shared tasks, such as Hope 2024 \cite{garcia-baena_overview_2024} and PolyHope \cite{butt_overview_2025}, have explored hope speech detection in English and Spanish across binary and multi-class classification conditions. Some common challenges observed across the systems include class imbalance, code-mixing and script-switching, and sociolinguistic differences \cite{butt_overview_2025}.

    It is clear from existing literature that hope speech detection is treated as a text classification problem \cite{kowsari_text_2019}. With a focus on optimizing model architectures, \citet{sharma_stop_2025} found that an ensemble approach offered the best model performance by combining the architectures of Long Short-Term Memory (LSTM; \citealt{hochreiter_long_1997}), mBERT \cite{devlin_bert_2019}, and XLM-RoBERTa \cite{conneau_unsupervised_2020} achieving an impressive weighted average $F_1$-score of 0.93 for English. However, solely focusing on model architectures fails to account for linguistic contexts beyond lexical structures \cite{li_text_2022}. We are now only observing approaches that account for the linguistic context, such as translation-based techniques \cite{ahmad_multilingual_2025}, but research within this area remain limited.
    
\section{Methodology}
\label{sec:methodology}

    With approximately 2.5 million speakers, Tulu is an under-resourced language \cite{shetty_natural_2024}. In contrast to other Dravidian languages, such as Kannada, Malayalam, and Tamil, resources for natural language processing remain scarce. Therefore, the primary purpose of \textit{Hope Speech Detection in Code-Mixed Tulu Language} shared task was to develop multi-level hope speech classification systems in code-mixed Tulu. We treat this shared task as a classification problem \cite{kowsari_text_2019}. We used XLM-RoBERTa as the base pre-trained language model (PLM) for our system \cite{conneau_unsupervised_2020}. To summarize, XLM-RoBERTa embeddings were trained on 2.5 TB of filtered web-crawled data containing 100 languages. However, it is important to note that Tulu was not one of these languages. Encoder-decoder transformer-based \textsc{PLMs} were used because of their ability to support domain adaptation, meaning that we can continue training a pretrained language model on additional in-domain data without the need to train a new PLM from scratch. 

    The shared task was broken down into two sub-tasks which we refer to as Task 1 and Task 2 \cite{thenmozhi_findings_2026}. For Task 1, the goal was to classify each comment into one of four coarse-grained categories: \textit{encouraging}, \textit{discouraging}, \textit{uninvolved}, and \textit{blended tone}. For Task 2, the goal was to classify hope-related content into five fine-grained categories: \textit{optimistic hope}, \textit{realistic hope}, \textit{inspiring hope}, \textit{fading hope}, and \textit{hopelessness}. For each task, we first developed a baseline system and then developed organically adapted candidate systems by further training XLM-RoBERTa on organically collected Tulu social media text containing code-mixed and mixed-script variation before downstream fine-tuning on the labeled shared-task data. For submission, we selected the best performing candidate system in each task based on development set macro average $F_1$-scores.

\subsection{Training Data}
\label{subsec:data}

    \begin{table}
        \caption{\label{tab:cg_train_labels} Coarse-Grained Training Class Labels}
        \centering
        \begin{tabularx}{\linewidth}{l*{1}{>{\centering\arraybackslash}X}}
            \toprule
            \textbf{Class} & \textbf{Count} \\
            \midrule
            Blended Tone & 895 \\
            Discouraging & 711 \\
            Encouraging & 1895 \\
            Uninvolved & 2490 \\
            \bottomrule
        \end{tabularx}
    \end{table}

    \begin{table}
        \caption{\label{tab:fg_train_labels} Fine-Grained Training Class Labels}
        \centering
        \begin{tabularx}{\linewidth}{l*{1}{>{\centering\arraybackslash}X}}
            \toprule
            \textbf{Class} & \textbf{Count} \\
            \midrule
            Fading Hope & 236 \\
            Hopelessness & 937 \\
            Inspiring Hope & 1129 \\
            Optimistic Hope & 380 \\
            Realistic Hope & 503 \\
            \bottomrule
        \end{tabularx}
    \end{table}
    
	The organizers of the \textit{Hope Speech Detection in Code-Mixed Tulu} shared task provided labeled training, development data, and a test set consisting of social media comments in code-mixed Tulu \cite{thenmozhi_findings_2026}. The training data was split into coarse-grained hope tone classification (Task 1) and fine-grained hope type classification (Task 2). These datasets formed the basis of our baseline system development and evaluation. The labeled training data was used to train the classification models, the development set was used for validation and model comparison, and the official test set was used only for final shared-task submission and evaluation.
    
	The distribution of target class labels in the training data is shown in Table \ref{tab:cg_train_labels} for the coarse-grained task and Table \ref{tab:fg_train_labels} for the fine-grained task. There is noticeable class imbalance across labels in both tracks. For the coarse-grained task, \textit{uninvolved} and \textit{encouraging} account for most of the training examples, while \textit{discouraging} and \textit{blended tone} appear less frequently. For the fine-grained task, \textit{inspiring hope} and \textit{hopelessness} are the most common labels, while \textit{fading hope}, \textit{optimistic hope}, and \textit{realistic hope} are less represented. 

\subsection{Domain Adaptation}
\label{subsec:domain_adaptation}

    The first stage in developing our system was domain adaptation, where we further adapted a PLM using organically collected Tulu social media text containing code-mixed and mixed-script variation before downstream supervised classification. \citet{liu_deep_2022} showed that domain adaptation can improve the performance of transformer-based language models in downstream tasks. We incorporated additional organic Tulu social media text for domain adaptation. This corpus consisted of naturally occurring Tulu social media comments and was intended to better reflect real-world code-mixed and mixed-script usage.
    
    Unlike the labeled shared-task comments, this organic data consisted of naturally occurring Tulu social media comments collected to better reflect the linguistic variation found in real-world usage, including mixed-script and code-mixed forms. Following the data development approach described in \citet{wong_cantnlplt-edi-2024_2024}, the organic Tulu data was retrieved using a Tulu language identification pipeline trained with Wikipedia-based data, which was used to identify candidate Tulu social media text for continued language-model adaptation. We used this organically collected Tulu social media text from the \textit{Global Corpus of Language Use} \cite{dunn_mapping_2020}, which contains code-mixed and mixed-script variation, as unlabeled in-domain data for further adapting XLM-RoBERTa before downstream classification.

    We adapted XLM-RoBERTa using a masked language modeling objective and then trained an \texttt{AutoModelForMaskedLM} model on this corpus. We trained the masked language model for two epochs with a per-device batch size of 16 and a learning rate of $5 \times 10^{-5}$. We used the AdamW optimization setup provided in the training framework \cite{loshchilov_decoupled_2019}. After training, we saved the adapted checkpoint and tokenizer for use in downstream supervised fine-tuning.

\subsection{Supervised Classification}
\label{subsec:supervised_classification}

    As discussed in Section \ref{subsec:domain_adaptation}, we developed our classification systems using both the original XLM-RoBERTa checkpoint and the organically adapted checkpoint produced during the domain adaptation stage. We then fine-tuned these models on the labeled shared-task data for the two Tulu classification settings: Task 1 (coarse-grained hope tone classification); and, Task 2  (fine-grained hope type classification). All model training was conducted in a shared Google Colab environment. Note that the relevant code notebook and scripts are available on GitHub\footnote{https://github.com/landrewi/cantnlp-HopeSpeechDetection-DravidianLangTech2026}. For each task, we trained a sequence classification model using the appropriate number of output labels. We used \texttt{AutoModelForSequenceClassification} with the XLM-RoBERTa tokenizer and then trained the classifier for up to four epochs with a learning rate of $2\times10^{-5}$, using the same fine-tuning configuration for both the baseline and organically adapted systems. Four candidate systems overall were produced: a baseline and an organically adapted model for both Task 1 and Task 2 each. The performance of these candidate systems on the development set is reported in Section \ref{sec:results}. Based on the development set macro average $F_1$-scores, we selected the organically adapted models as our submitted systems for both Task 1 and Task 2.

\section{Results}
\label{sec:results}

    The macro average $F_1$-scores of our candidate models on the development set are presented in Table \ref{tab:cg_validation_results} and Table \ref{tab:fg_validation_results}. Both organic models performed better than the baseline models on the development set. In Task 1, the best performing candidate model was the organic model which yielded a macro average $F_1$-score of 0.5238 compared to 0.5227 for the baseline model. In Task 2, the best performing candidate model was also the organic model which yielded a macro average $F_1$-score of 0.3416 compared to 0.3171 for the baseline model. Based on these model performance metrics, we therefore selected the organically adapted models as our submitted systems for both Task 1 and Task 2.

    \begin{table}[t]
        \caption{\label{tab:cg_validation_results} Validation results for Task 1: Coarse-Grained Hope Speech Classification. The best performing model based on macro average $F_1$-score is in \textbf{bold}.}
        \centering
        \begin{tabularx}{\linewidth}{l*{2}{>{\centering\arraybackslash}X}}
            \toprule
            \textbf{Metric} & \textbf{Baseline} & \textbf{Organic} \\
            \midrule
            Accuracy & 0.6269 & 0.6869 \\
            Macro $F_1$ & 0.5227 & \textbf{0.5238} \\
            Weighted $F_1$ & 0.6364 & 0.6545 \\
            \bottomrule
        \end{tabularx}
    \end{table}

    \begin{table}[t]
        \caption{\label{tab:fg_validation_results} Validation results for Task 2: Fine-Grained Hope Speech Classification. The best performing model based on macro average $F_1$-score is in \textbf{bold}.}
        \centering
        \begin{tabularx}{\linewidth}{l*{2}{>{\centering\arraybackslash}X}}
            \toprule
            \textbf{Metric} & \textbf{Baseline} & \textbf{Organic} \\
            \midrule
            Accuracy & 0.4135 & 0.5337 \\
            Macro $F_1$ & 0.3171 & \textbf{0.3416} \\
            Weighted $F_1$ & 0.3753 & 0.4732 \\
            \bottomrule
        \end{tabularx}
    \end{table}

    \begin{table}[t]
        \caption{\label{tab:test_evaluation} Macro average $F_1$-score from official test evaluation and rank for both tasks.}
        \centering
        \begin{tabularx}{\linewidth}{l*{2}{>{\centering\arraybackslash}X}}
            \toprule
            \textbf{Task} & \textbf{$F_1$-score} & \textbf{Rank} \\
            \midrule
            Coarse-Grained & 0.50 & 5 \\
            Fine-Grained & 0.33 & 7 \\
            \bottomrule
        \end{tabularx}
    \end{table}

\section{Discussion}
\label{sec:discussion}

    Based on the evaluation metrics alone, our organically adapted systems performed modestly better than the baseline systems on the development set for both tasks. With reference to Table \ref{tab:cg_validation_results} and Table \ref{tab:fg_validation_results}, we see that the organically adapted model improved model performance over the baseline model in both the coarse-grained and fine-grained conditions. In Task 1, the macro average $F_1$-score increased from 0.5227 to 0.5238, while in Task 2 the macro average $F_1$-score increased from 0.3171 to 0.3416, suggesting that further adapting XLM-RoBERTa on organically collected Tulu social media text containing code-mixed and mixed-script variation provided a measurable benefit for hope speech classification, particularly for the fine-grained task (Task 2).

    While the development set results indicate consistent improvement, they were not uniform across metrics and tasks. In Task 1, the macro average $F_1$ improvement was small, but accuracy and weighted $F_1$ increased more clearly, possibly reflecting that the organically adapted model better captured the majority-class patterns while still struggling with minority-class distinctions. As for Task 2, the organically adapted model improved across all reported metrics, suggesting that exposing the model to organically collected in-domain text containing code-mixed and mixed-script variation may be specifically helpful when the label space was larger and the distinctions between classes were more subtle.

    When comparing our submitted systems with the official test evaluation results found in Table \ref{tab:test_evaluation}, our organically adapted systems achieved a macro average $F_1$-score of 0.50 for Task 1 and 0.33 for Task 2, ranking fifth and seventh overall, respectively. The best performing team in Task 1 was Team \texttt{CUET\_Synthetica} which yielded a macro average $F_1$-score of 0.58, while we yielded a macro average $F_1$-score of 0.50 and ranked 5th overall. The best performing team in Task 2 was also Team \texttt{CUET\_Synthetica} which yielded a macro average $F_1$-score of 0.42, while we yielded a macro average $F_1$-score of 0.33 and ranked 7th overall. Despite this mid-ranged performance, these results indicate that the approach of adapting multilingual pretrained models using organically collected unlabeled Tulu social media text for further language-model adaptation is viable in this low-resource code-mixed setting, but the complexity of code-mixed Tulu requires more robust data or architectural innovations.

\section{Conclusion}
\label{sec:conclusion}

    While our submitted systems did not achieve the best results, our organically adapted approach produced improvements over the baseline models on the development set for both Task 1 and Task 2. These results suggest that further adapting XLM-RoBERTa on organically collected Tulu social media text containing code-mixed and mixed-script variation can improve hope speech detection in code-mixed Tulu. At the same time, the official test results indicate that this task remains challenging, especially under class imbalance and fine-grained label distinctions.

\section*{Limitations}
\label{sec:limitations}

    One limitation of the current study is that we only evaluated one domain adaptation strategy based on organically collected Tulu social media text and did not compare it directly with a synthetic script-switching variant such as that explored in \cite{wong_cantnlplt-edi-2024_2024}. This prior work explored both synthetic and organic script-switching for domain adaptation, whereas our current system focused only on further adapting XLM-RoBERTa using organically collected Tulu social media text containing code-mixed and mixed-script variation. As a result, we cannot determine whether the gains observed on the development set were specific to the organic adaptation pipeline or whether a synthetic approach would have produced similar improvements.

    Another limitation was class imbalance in both tasks, as shown in Table \ref{tab:cg_train_labels} and Table \ref{tab:fg_train_labels}. In Task 1, the training set was dominated by the \textit{uninvolved} and \textit{encouraging} classes, while \textit{discouraging} and \textit{blended tone} were underrepresented. In Task 2, \textit{inspiring hope} and \textit{hopelessness} appeared most frequently, while \textit{fading hope}, \textit{optimistic hope}, and \textit{realistic hope} were underrepresented. We did not apply any explicit class-imbalance mitigation techniques such as class-weighted loss, focal loss, or over/under-sampling during training. Future work could test whether these strategies improve performance on minority classes, particularly for Task 2. This imbalance may reduce the model’s ability to learn reliable distinctions for less frequent hope-related categories and may have contributed to the modest official test performance.
    
    Finally, beyond the upstream and downstream impacts of bias in multilingual \textsc{PLMs}, we also recognize that incorporating externally collected digital language data for language-model adaptation may introduce additional biases not fully addressed in this paper, such as sampling bias, geographic bias, and variation tied to platform-specific language-use (\citealp{goldhahn_building_2012}; \citealp{dunn_mapping_2020}). These limitations suggest that future work should compare multiple adaptation strategies more directly and further examine the effects of data balance and external-data bias in code-mixed Tulu hope speech detection.

\section*{Ethical Considerations}
\label{sec:ethical_considerations}

    Our system contributes to much needed research on under-resourced languages such as Tulu \cite{shetty_natural_2024}. The results suggest that existing PLMs, such as XLM-RoBERTa \cite{conneau_unsupervised_2020}, are ill-equipped to meet the needs of Tulu language speakers. Therefore, the findings from our paper provides offers an insightful benchmark for Tulu within the area of hope speech detection. However, we need to think beyond development, but also applied to benefit Tulu speaking communities. Much like hate speech detection \cite{wong_what_2024}, we should also adopt a similar view to determine the social impact of hope speech detection research. Moreover, we should consider how systems developed can be deployed to detect hope speech in real-world social media contexts.
    

\bibliography{references}



\end{document}